\documentclass[letterpaper, 10 pt, conference]{ieeeconf}  % Comment this line out if you need a4paper

\usepackage{comment}
\usepackage{graphicx}
\usepackage{subcaption}
\usepackage{color}
\usepackage{xfrac}
\usepackage{array}
\usepackage{gensymb}
\usepackage{hyperref}
\usepackage{soul}
\usepackage{amsmath}
\usepackage{tabularx,ragged2e,booktabs,caption}
\usepackage[textsize=tiny]{todonotes}
\usepackage[detect-all,separate-uncertainty=true,multi-part-units=single]{siunitx}

\newcommand{\rev}[1]{{\color{black} #1}} % red

\IEEEoverridecommandlockouts                              % This command is only needed if 
\title{\LARGE \bf
The More the Merrier: Running Multiple Neuromorphic Components On-Chip for Robotic Control
}
\author{Evan Eames$^{a,\dagger}$, Priyadarshini Kannan$^{b}$, Ronan Sangouard$^{b}$, Philipp Plank$^{c}$, Elvin Hajizada$^{c}$, Gintautas\\ Palinauskas$^{b}$, Lana Amaya$^{b}$, Michael Neumeier$^{a,d,\dagger}$, Sai Thejeshwar Sharma$^{d}$, Marcella Toth$^{d}$,\\ Prottush Sarkar$^{d}$, Axel von Arnim$^{b}$
\thanks{$^{a}$~The author was with the Department of Neuromorphic Computing, fortiss -- Research Institute of the Free State of Bavaria, Munich, Germany. He is now at the Ludwig Maximillian University of Munich
        }
\thanks{$^{b}$~Department of Neuromorphic Computing, fortiss -- Research Institute of the Free State of Bavaria, Munich, Germany
        }
\thanks{$^{c}$~Intel Neuromorphic Computing Lab, Munich, Germany
        }
\thanks{$^{d}$~Technical University of Munich (TUM), Munich, Germany
        }
\thanks{\textdagger~Corresponding Author:}
\thanks{~~~~~~~evan.eames@lmu.de}
\thanks{~~~~~~~michael1.neumeier@tum.de}
}

\begin{document}

\maketitle
\thispagestyle{empty}
\pagestyle{empty}

%%%%%%%%%%%%%%%%%%%%%%%%%%%%%%%%%%%%%%%%%%%%%%%%%%%%%%%%%%%%%%%%%%%%%%%%%%%%%%%%
\begin{abstract}

It has long been realized that neuromorphic hardware offers benefits for the domain of robotics such as low energy, low latency, as well as unique methods of learning. In aiming for more complex tasks, especially those incorporating multimodal data, one hurdle continuing to prevent their realization is an inability to orchestrate multiple networks on neuromorphic hardware without resorting to off-chip process management logic. To address this, we show a first example of a pipeline for vision-based robot control in which numerous complex networks can be run entirely on hardware via the use of a spiking neural state machine for process orchestration. The pipeline is validated on the Intel Loihi 2 research chip. We show that all components can run concurrently on-chip in the milli Watt regime at latencies competitive with the state-of-the-art. An equivalent network on simulated hardware is shown to accomplish robotic arm plug insertion in simulation, and the core elements of the pipeline are additionally tested on a real robotic arm.

\end{abstract}
\IEEEpeerreviewmaketitle

%%%%%%%%%%%%%%%%%%%%%%%%%%%%%%%%%%%%%%%%%%%%%%%%%%%%%%%%%%%%%%%%%%%%%%%%%%%%%%%%
\section{INTRODUCTION}
\label{sec:Introduction}

As the domain matures, neuromorphic chips running spiking neural networks (SNN) have seen increased robotic integration. These ``neurorobots'' have accomplished real-time drone control \cite{Paredes-Valles24,Vitale21}, robotic insertion \cite{Amaya23,Amaya24}, autonomous navigation \cite{Dumont23,Jones20,Tang19,Kreiser18b}, object avoidance \cite{Jiang25,Mitchell17}, among numerous other examples.

This first generation of neurorobots all share one common trait: they accomplish these tasks through deploying singular SNNs. This has served as testament to the versatility of the SNN. In pushing towards achieving more advanced robotic behavior, it is entirely reasonable to continue deploying larger and deeper SNNs. Yet, one may instead ask whether using multiple spiking components could be more productive. Indeed, evolution has endowed biological systems with uncountable different forms of neuronal structure, all specialized in form and function toward optimizing individual goals, yet all unified in supporting the survival of the sentient being within which they operate.

In this direction, perhaps applying this inspiration to neurorobots could extend the above-mentioned examples beyond the initial steps they represent to robust human-like systems.

The current status quo when connecting multiple SNN has been to combine the neuromorphic hardware running individual SNNs with traditional non-spiking architecture for overarching organization \cite{Christensen22,Bartolozzi22}. Yet this voids many of the intrinsic advantages of neuromorphic hardware (i.e., low-latency and energy efficiency). \cite{Liang19} introduces a method for robot control in which segments of an SNN can be activated or suppressed on-chip via arbitration neurons: a promising first step, though not yet generalized to multiple heterogeneous spiking components.

Yet another drawback to this state of affairs is the lack of ability to combine multisensorial information --- an interplay in biological systems responsible for painting the picture of the reality we inhabit. While neuromorphic chips have been applied to vision (survey: \cite{Gallego22}), touch (survey: \cite{Liu25}), sound (survey: \cite{Basu25}), and even smell (ie. \cite{Dennler25}), the inability to integrate multiple networks on-chip results in these efforts remaining isolated \cite{Bartolozzi22}. Neurorobotic grasping, for example, although sufficiently developed in itself \cite{Volinski21, Ehrlich22,Li25,Sankar25}, cannot currently exist within a purely neuromorphic robot system (ie. combined with a neuromorphic arm or neuromorphic vision sensor).

Within the realm of robotic vision, a substantial effort has gone into exploring biologically inspired on-chip attention and saliency algorithms \cite{Ghosh22,D’Angelo22,Molin21,Kreiser20,Subramaniam17,Adams14}. Yet again, although these results are promising, there is no obvious way to extend them. A system through which neuronal fields handling vision data can be subsequently wired to downstream SNNs for robot control has remained elusive.

The result of this fundamental absence of unifying neuromorphic orchestrational logic --- that which would allow for multi-component pipelines to run on-chip --- is neurorobotics remaining fractured and immature as a domain.

Attempts to reconcile the situation through understanding the role of spiking systems in more complex behaviors have remained largely confined to the realms of neuroscience and biology \cite{Dayan08,Panzeri17}. Others have recognized the need to incorporate multimodal data as is characteristic of modern non-neuromorphic robots \cite{Putra24,Bhanja23} for which a path forward has not been found. The recent emergence of novel neuromorphic platforms which, in theory, could support multiple components and multi-sensor processing \cite{Mayr19,Shrestha23,innatera_pulsar} further motivates this imperative.

To address this, we present a first example of multiple neuromorphic components all interconnected and running concurrently within neuromorphic hardware, without the need for off-chip traditional logic for process orchestration.

As the motivation of this pursuit is neurorobots, we develop the pipeline within the context of a robot arm attempting to insert multiple plugs into their corresponding sockets. The network involves a large number of neuronal elements: an attention method implemented using a dynamic neural field, a separate memory field to retain the locations of previously searched sockets, a gating mechanism to focus on a specific region of the field of view, a classifier, a logical system to check if we have found the desired socket, a visual servoing actor, and an overarching neural state machine for process orchestration. All of these run on-chip, on a single board, without requiring off-chip components for process orchestration. An identical network architecture is tested on a simulated robotic arm, and individual components are additionally tested on a real robot.

We assert that the following are novel:

\begin{itemize}
\item A multi-component spiking pipeline run on neuromorphic hardware (in the sub-milli Watt regime) without requiring off-chip logic for process management
\item A hardware implementation of a Neural State Machine in which states and state-changes are mediated via Dynamic Neural Field
\end{itemize}

In addition, construction of this pipeline resulted in a novel implementation of dynamically gating a neuronal field on-chip at run time as well as a fully neuron-based visual servoing actor.

\section{THEORETICAL BACKGROUND}

\subsection{Dynamic Vision Sensor}
\label{sec:DVS_Intro}

A dynamic vision sensor (DVS) is a type of neuromorphic camera. As opposed to pixel-specific RGB values in traditional cameras, in a DVS photoreceptors emit timestamped energy pules (known as `events') for brightness changes, thus avoiding redundant information \cite{Gallego22}. Along with biologically inspired spiking neural networks (SNNs), such technologies are seeing growing adoption on account of their energy savings as well as high dynamic ranges, and are considered ideal for edge devices. Applications include drone control \cite{Paredes-Valles24}, visual servoing \cite{Ayyad23,Muthusamy21,Lawson23}, space-based situation awareness \cite{Saeed20,Ralph22}, and robotic navigation \cite{Mitchell17}, among others.

\subsection{Dynamic Neural Field}
\label{sec:DNF_Intro}

A dynamic neural field (DNF) \cite{Schorner15} is an attention mechanism allowing us to focus on one or multiple regions of interest within a FoV. These mimic the interactions and activations of neuronal populations, and allow for localized activity ``peaks" within a population. Each peak represents a region of interest to be focused on. DNFs have since been extended to various topics such as trajectory planning and robot control \cite{Erlhagen06,Sandamirskaya13b,Martel16}, among others.

The DNF equation is an activation function defined over a feature dimension such as color, retinal location, orientation, or direction of movement. It represents the dynamics of the activations of a group of neurons interacting with one another. The equation is given as:
\begin{multline}
    \tau\dot{u}(x,t) = -u(x,t) + h + S(x,t)\\ + \int{|x-x'|g(u(x',t))\delta x'}
\label{eq:DNF}
\end{multline}
\noindent where $u(x,t)$ is DNF activity for a feature $x$ at a time $t$, $h$ is the negative resting level which prevents output in the absence of external input ($S$), and the integral represents the interaction kernel whose function $g$ could be Gaussian or Difference-of-Gaussian facilitating inhibitory and excitatory interactions with other neurons $x'$ of the DNF \cite{Schorner15}.

If the kernel is set as a Ricker wavelet \cite{Ricker53} (in 2 dimensions referred to as a ``Mexican hat'') the DNF is said to be ``multipeak''. Multiple peaks can form and be sustained, yet if any two come too close to one another then interference will cause one or both to collapse. If the kernel is set to be Gaussian (with a negative baseline) then the DNF is said to be ``selective''. The strongest peak will collapse all others through global inhibition.

DNF have been shown to interface especially well with tasks involving vision data (saliency \cite{Sandamirskaya14}, tracking \cite{Martel16}, etc), and to represent an intrinsic form of memory \cite{Johnson09,Lozenski22}.

\subsection{Neural State Machine}
\label{sec:NSM_Intro}

Neural state machines (NSMs) aim to extend multi-process organizational logic into spiking architecture. They allow for multiple neuronal processes to be executed concurrently or sequentially on neuromorphic hardware without the need to resort to classic off-chip logic gates in managing said processes.

Investigations of NSMs have  generally been built around winner-take-all (WTA) layers as mediating state changes \cite{Neftci13,Liang17,Kreiser18a,Liang19}. These have thus far been applied to relatively simple tasks in which states can be represented by a single neuron (ie. heart rate patterns \cite{Carpegna24}, direction of movement \cite{Liang19}, time durations to stay in a state \cite{Cotteret25}, etc.).

A selective DNF is effectively a biologically inspired WTA implementation. In the context of NSMs a given DNF activation pattern may correspond to a state --- paired with an arbitration systems of neurons which can suppress the field's activation when a condition is satisfied, hence instigating a state change \cite{Sandamirskaya10,Sandamirskaya11,Richter18}. Much of the theory can be found in \cite{Richter18} (section 3.5), as well as in \cite{Sandamirskaya10,Sandamirskaya11}. Beyond the theory, in the 2 dimensional case this avenue involves substantially higher neurons and synapses on account of the kernels, and has thus far not been successfully implemented on real hardware.

Within an NSM, complex behavior is divided into modular ``\emph{Elementary Behaviors}'' (EBs). These can be unitary objectives such as ``search for and classify socket'', ``descend towards socket'', ``align socket'', etc. which contribute to a larger task (ie. ``find socket and insert plug''). Each EB is mediated by arbitration neuronal structure (which, case dependent, can be a single neuron, field of neurons, on something more abstract). These include:\\

\noindent \emph{i. Intention}

Each EB possesses, and is initialized by, an \emph{intention} node. When a task is initiated the intentions of all required EB are activated. Until the EB is completed, the intention node will remain activated. The intention node can also affect sensory-motor systems, attuning these as applicable for a given EB.\\

\noindent \emph{ii. Condition of Satisfaction}

Each EB also possesses a Condition of Satisfaction (CoS). When the intention node is activated it subsequently activates the CoS, which in turn begins waiting for some expected input representing the EB to be completed. When this is detected, the CoS is excited beyond some threshold, at which point it both inhibits the intention of the current EB and simultaneously activates the intentions of any downstream EBs (see \emph{Precondition} below).\\

\noindent \emph{iii. Condition of Dissatisfaction}

The CoD works much the same as the CoS, however the signal it awaits is that indicating that a goal has not been satisfied (ie. the incorrect socket has been found). This may inhibit the intention and trigger a downstream corrective EB, or it may leave the intention activated and instigate a repetition within the current EB.\\

\noindent \emph{iv. Precondition}

Precondition nodes link multiple EBs that must be completed sequentially. A precondition node inhibits any downstream EB intention nodes. Only when a CoS inhibits the precondition node are these subsequently `released' and hence downstream EBs can initiate.

\section{ARCHITECTURE DESCRIPTION}
\label{sec:Architecture}

Even within traditional robotics, plug insertion remains a challenging problem as well as an area of active research \cite{Cressman23,Hartisch23,Spector21,Vecerik18}. The difficulty comes from the multifaceted nature of the problem: the correct socket must first be identified, which involves vision data incorporation; the approach must be especially precise; the final alignment must be delicate enough to avoid damaging the plug; etc. For these reasons the task acts as a perfect testbed for multi-component neuromorphic architectures of complexities not yet explored.

\begin{figure*}[h]
\centering
\includegraphics[width=\textwidth]{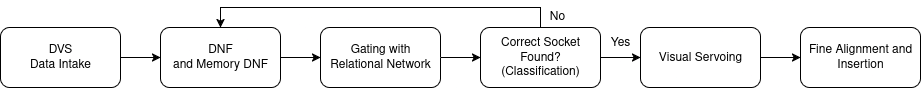}
\caption{Pipeline flow diagram. A full architecture with the NSM components is included in Figure \ref{fig:full_architecture}.}
\label{fig:flow_diagram}
\end{figure*}

To attempt plug insertion with a neurorobot we design a neuromorphic pipeline consisting of six primary components (see Figure \ref{fig:flow_diagram}) which are here enumerated. These are subsequently combined into a non-linear NSM (section \ref{sec:Full Architecture} and Figure \ref{fig:full_architecture}).

The pipeline is built in LAVA\footnote{https://lava-nc.org/} version 0.10.0 and run on a Loihi 2 Kapoho Point board (Oheo Gulch) running within the Intel Neuromorphic Research Community Vlab cloud \cite{Orchard21}. Neurons are based on the Leaky Integrate-and-Fire (LIF) model \cite{Abbott99}, abstracted in LAVA as:\\
\begin{equation}
\label{eq:u}
u(t) = u(t-1)(1-du) + a_{in}
\end{equation}
\begin{align}
\label{eq:v}
v(t) =
\left\{
\begin{aligned}
& v(t-1)(1-dv) + u(t) + b & \textrm{if } & v(t-1) < v_{th} \\
& 0 & \textrm{if } & v(t-1) \geq v_{th}
\end{aligned}
\right.
\end{align}
where $u$ and $v$ are the neuron's current and voltage, respectively, with $du$ and $dv$ the inverse of their decay time-constants, $a_{in}$ is input from other neurons, $b$ is a bias term (herein zero unless otherwise indicated) and $v_{th}$ is the voltage threshold above which a spike is sent onward and the voltage is reset.

The pipeline also includes two minor variants of the LIF neurons (whose usage will be explicitly mentioned for relevant pipeline segments). These are ``LIF Reset'' neurons where the voltage is reset to zero after each timestep regardless of whether the neuron has fired, and ``LIF Refractory\footnote{In LAVA there is not yet an on-chip implementation of the LIF Refractory neuron, however the ``Hard Coded LIF'' (NcModelLifHC) also allows us to set a refractory delay.}'' neurons where the voltage remains zero for a predetermined number of timesteps after firing.

\subsection{Data Capture}
\label{sec:Data_Capture}
We opt for the eye-in-hand configuration as it allows the camera to be brought close to the sockets in order to see them in better detail. More information on the camera mounting can be found in section \ref{sec:sim} (see also Figure \ref{fig:camera_setup}). As DVS cameras require relative motion for event generation, we attach a vibrational device to the arm-mounted-camera\footnote{We use a small bullet vibrator.} to imitate ocular microtremors (OMTs) \cite{Alexander19}. The frequency of vibrations is roughly 100~Hz, consistent with OMTs in the human eye \cite{Bolger99}. Indeed, the resulting images are clearer than with microsaccades or drift motion, with well defined edges. OMTs have the additional benefit that movement of the sockets in the FoV is negligible.

Events are binned into 20 ms frames. Polarities are not used. On account of I/O constraints with the current generation of neuromorphic hardware, we additionally downsample the resolution from $640\times480$ to $80\times80$ before it is fed into the pipeline (using a single average pooling layer). Event data is loaded and then sent to an $80\times80$ LIF neuron population as spikes via a LAVA Injector and PyToNxAdapter, the latter acting as an intermediary for binary-to-spike conversion. Connections are all sparse.

\subsection{Selective DNF}
\label{sec:DNF}
To focus on a single salient region we apply a DNF (section \ref{sec:DNF_Intro}). The DNF is chosen over other methods (ie. center-of event mass) as it is less susceptible to noise and will remain stabilized even if the selected object is momentarily obscured \cite{Sandamirskaya14}, as well as the benefits described in section \ref{sec:Introduction}.

For our application, we implement the DNF as an $80\times80$ field of LIF neurons with a single peak (WTA) Gaussian interaction kernel with event density (within the area of the peak) as the feature $x$ in equation \ref{eq:DNF}. This results in a peak, representing a salient region, and is seen within the bottom-left grid in Figure \ref{fig:RN_arch}. Henceforth this is referred to as the ``Selective DNF''.

In LAVA the built-in selective DNF kernel implementation invokes all-to-all neuron connectivity for global inhibition. In our case this is found to be too expensive ($80^2\times80^2 \approx 4\cdot10^7$ connections). We therefore set this built-in kernel to have zero inhibitory amplitude while instead using a single intermediary LIF Reset neuron. The intermediary neuron is connected such that it receives the output of all neurons in the selective DNF field and, upon spiking, inhibits all neurons in the field. All connections are sparse. The intermediate neuron has $v_{th}$ set to be 10 times larger than the current it receives from a single LIF neuron. This allows for a small (10 neuron) peak to begin forming before global inhibition starts. At lower values of $v_{th}$ the peak is not large enough to easily self-sustain when the inhibition begins, and it collapses.

Corresponding parameters can be found in appendix \ref{sec:Parameters}.

\subsection{Gating with Relational Network}
\label{sec:RN}

The DNF allows us to select one of the four sockets for further investigation, however we still need to gate it for classification. This is somewhat analogous to cropping in classical computer vision, and ``covert attention'' in biological systems. A dynamic system for gating different regions of a 2D neuronal field during runtime is non-trivial. A partial solution is implemented in \cite{Xu24} for on-chip convolution, however it involves substantial non-neural computational overhead in order to arrange events spatially based on chronological order of arrival, and subsequently is incompatible with both stateful neurons and recurrent connections.

In the 1D case a similar task has been accomplished via a relational network\footnote{Called a ``Relational Field" in some publications.} (RN) \cite{Diehl16,Zhao20,Richter21}. Two 1D groups of neurons are connected to a third 2D group, representing all possible combinations between the neurons of the initial two groups. For a 2D field, to avoid the large number of neurons required for what would be a 4D combination group (roughly 40 million neurons as was the case with the DNF global inhibition), we instead combine regions of the FoV, as opposed to individual neurons. As our setup guarantees the FoV will contain one socket in each quadrant, we define four zones of interest each containing $40\times40$ neurons and design the RN around these.

\begin{figure*}[h]
\centering
\includegraphics[width=\textwidth]{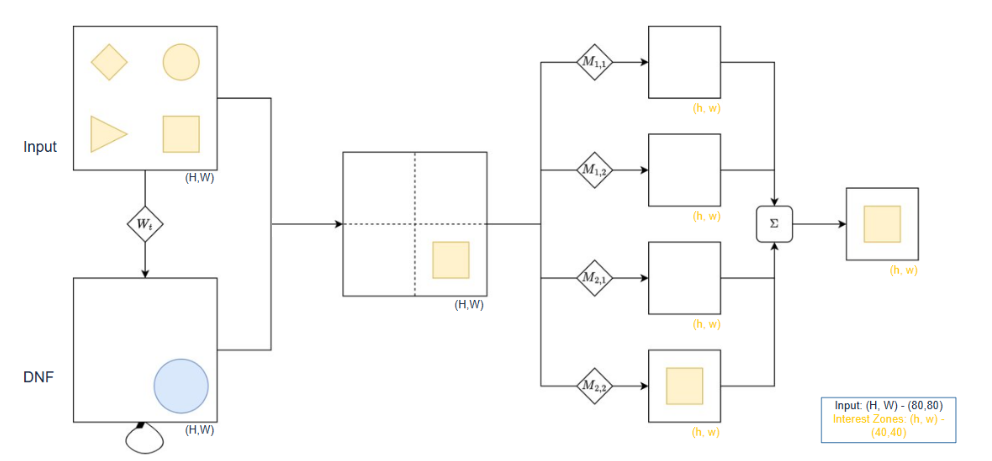}
\caption{RN Architecture. Each square represents a 2D field of neurons --- larger squares are $80\times80$, smaller squares are $80\times40$. The input creates a region of activity in the selective DNF (left). These are then both connected to a ``select neuron field" (center large square). Four weight matrices $M_{i,j}$ connect the quadrants to a single output LIF field (right), hence completing the gating. The Input and DNF contain LIF neurons, while the select neuron field and gated field are both LIF Reset.}
\label{fig:RN_arch}
\end{figure*}

Explicitly, the input and selective DNF are both connected to a new field of LIF Reset neurons with higher $v_{th}$ (referred to as the ``select neuron field'') such that these neurons only spike if they receive current from both the input as well as from the DNF at the same time. This has the effect that only the socket upon which the selective DNF peak has formed will appear in this new field. Dense connections then connect the quadrants of the select neuron field to a final $40\times40$ LIF Reset neuron field, effectively performing a summation across the quadrants. Recall that three of the select neuron field quadrants should not be spiking, and therefore the final ``gated'' field will contain only the socket of interest. The full RN architecture is shown in figure~\ref{fig:RN_arch}.

\subsection{Classification}
\label{sec:Classification}

Classification is accomplished through a spiking feedforward network (see architecture details in appendix \ref{sec:Parameters}) trained using the SLAYER method \cite{Shrestha2018}. As in \cite{Palinauskas23} the training was performed for 100 epochs with a batch size of 16, a learning rate of 0.001, and the ADAM optimizer. 

The output of the classifier is processed using two groups of 4 LIF Reset neurons, with each neuron representing one of the four socket classes. These are the ``matching circuit'' and ``non-matching circuit'' (see Figure \ref{fig:full_architecture}). Both of these are also connected to a predefined user input representing the desired socket (instantiated using the LAVA UserInput process). This enters the network using a PyToNxAdapter and is converted into spikes. For example, if the user input is set to [1,0,0,0] (where the first class corresponds to the USB) then the adapter will send spikes only to the first neuron of each 4-neuron circuit\footnote{There is slightly more nuance here. The non-matching circuit must receive the bitwise negation of the user input. That is to say, if the user input is [1,0,0,0], the non-matching circuit must receive [0,1,1,1]. This is accomplished by first setting a bias voltage for all 4 neurons in the non-matching circuit. The user input is then split, with excitatory connections to the matching circuit and inhibitory connections to the non-matching circuit. The 3 neurons not inhibited by the user input will spike when incoming spikes are received from the classifier.}. The $v_{th}$ for all circuit neurons is set to twice the weights of the incoming connections. As they are LIF Reset, they will only spike if the neuron representing a class receives spikes from both the classifier and user input at the same time. Both circuits have their spikes aggregated and sent onward to the CoS and CoD neurons (see section \ref{sec:Full Architecture}).

\subsection{Inhibition via DNF Memory Field}
\label{sec:Memory}

To avoid repeatedly classifying the incorrect socket, we wish to commit the location of any classified sockets to memory. This is made possible using a multipeak DNF (henceforth referred to as a memory DNF) to which the selective DNF is connected with excitatory connections (as explored in \cite{Johnson09,Lins14}). Unlike the selective DNF, the memory DNF excitation amplitude is set high enough that the activation peaks remain once formed. The memory DNF is then connected back to the selective DNF with inhibitory connections. This assures that activation peaks cannot re-form in regions already explored and consecrated to memory\footnote{In fact, activation peaks within the memory DNF are not alone enough to suppress corresponding peaks in the selective DNF. These peaks only collapse when the entire selective DNF is further inhibited by the CoD neuron (see section \ref{sec:Full Architecture}). At this point the inhibition from the corresponding memory DNF peak prevents the selective peak from forming on the same socket.}.

We use LAVA's built-in multipeak kernel. Unlike with the selective DNF, the multipeak kernel does not require expensive all-to-all connectivity, and therefore we do not need to invoke another intermediary neuron. The corresponding parameters can be found in appendix \ref{sec:Parameters}.

\subsection{Visual Servoing}
\label{sec:VS}

\begin{figure}[h]
\centering
\includegraphics[width=0.48\textwidth]{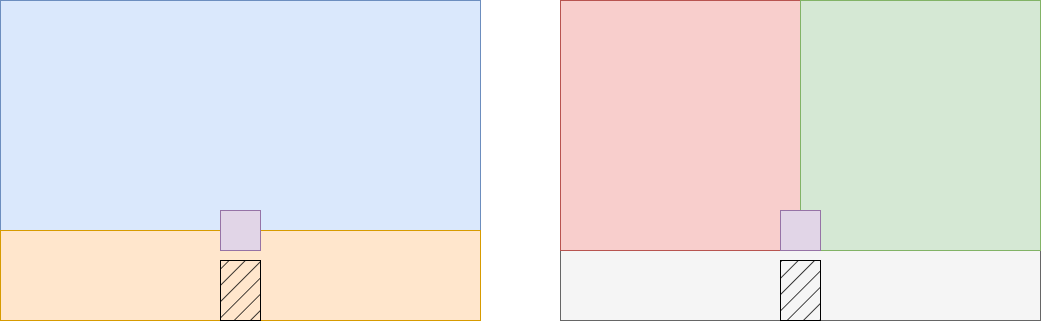}
\caption{For visual servoing the DNF is connected to two $80\times80$ directional fields in which regions correspond to directions of robot motion (these are in fact abstractions of the weight matrix connecting the selective DNF to the 5 directional neurons). Within these fields the colors represent the following end-effector translations: blue = $+y$, yellow = $-y$, red = $-x$, green = $+x$, purple = $-z$. The shaded region indicates where the tip of the end-effector peg occludes a region of the FoV in the general case (in our case we adjust the camera such that the occlusion is negligible -- see section \ref{sec:sim}).}
\label{fig:visual_servoing}
\end{figure}

Once we have identified the correct socket we wish to initiate visual servoing. The selective DNF can be used for this purpose as it will continue to track the desired socket's position within the FoV (similar to \cite{Martel16}). To translate this into robot motion we design a simplified spiking actor. The selective DNF is connected to an array of 5 neurons each representing a translation of the end-effector ($\pm x$, $\pm y$, and $-z$). The weights serve to quantify which region of the FoV the DNF peak is located, and can be abstracted as two directional fields. The first field guides the robot's motion along the $\pm x$ axis, while the second guides the motion along the $\pm y$ axis --- ie. if in the first field the DNF peak excites the neurons on the left then the Loihi chip will instruct the robot arm to move left. In both fields there is a small region which, if excited by the DNF peak, instructs the arm to move in the $-z$ direction (descending towards the socketboard). This is illustrated in Figure \ref{fig:visual_servoing}.

Recall that there is an offset between the end-effector peg and DVS camera (see Figure \ref{fig:camera_setup}). On account of this, when the plug is directly above the socket, movement in the $z$ direction has the effect of shifting the socket, and subsequently the DNF peak, in the FoV. In theory this results in a step-wise descent in which the peg descends incrementally, repeatedly re-aligning the socket in the $xy-\textrm{plane}$ FoV before resuming descent. However, in practice the offset is small enough that the descent is relatively smooth (discussed further in section \ref{sec:real})

\subsection{Insertion}
\label{sec:Insertion}

Once the plug makes contact with the table the fine alignment and insertion are accomplished using the force-torque actor outlined in \cite{Amaya24}. This has been trained using spiking RL, specifically with PopSAN \cite{Tang20}. The code is altered for a smaller search area: a Gaussian distribution centered on the point of plug impact the socketboard, with $\sigma = $ 4 mm.

Note that, on account of the I/O constraints mentioned in section \ref{sec:Data_Capture}, the downsampled resolution resulted in angular orientation control being infeasible. Therefore the angle of the plug and sockets are kept fixed, and only translational motion is controlled.

\subsection{Orchestrating Pipeline with NSM}
\label{sec:Full Architecture}

\begin{figure*}[h]
\centering
\includegraphics[width=\textwidth]{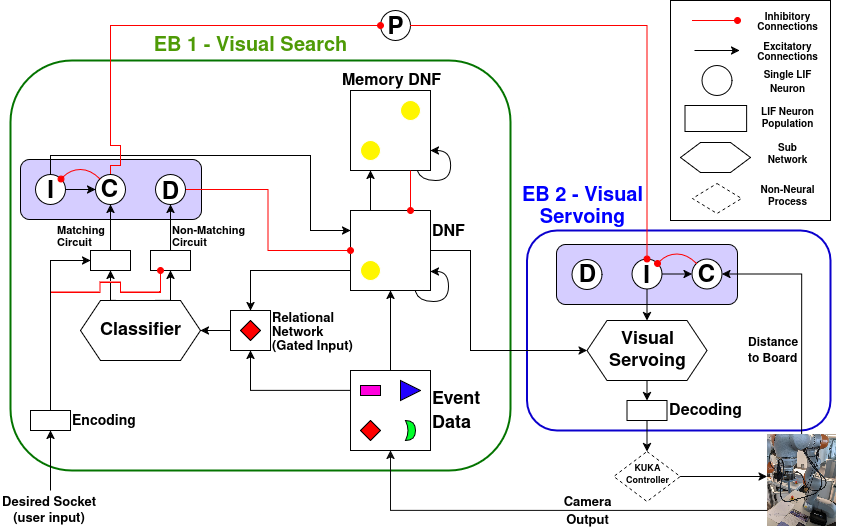}
\caption{Full pipeline architecture. The pipeline is divided into two elementary behaviors (EBs). The first comprises the search behavior and the second the visual servoing. Violet boxes represent the logic blocks guiding the NSM: I = Intention, C = Condition of Satisfaction, D = Condition of Dissatisfaction. Additionally, P is the Precondition node (which initiates the next EB).}
\label{fig:full_architecture}
\end{figure*}

Now that we have all the components, we organize and order the processes with the NSM elements listed in section \ref{sec:NSM_Intro}. Thus we arrive at the final architecture, which is presented in Figure \ref{fig:full_architecture}.\\

\noindent \emph{EB 1 -- Visual Search}

The first EB contains all elements related to the socket search. For our case the intention, CoS, and CoD (section \ref{sec:NSM_Intro}) are all single neurons (LIF, LIF Reset, and LIF Refractory respectively). The intention could be activated by some initialization system triggering the entire pipeline, however, for simplicity it is activated by default through the use of the LAVA LIF $bias$ parameter, and continually excites the DNF. The classifier output representing the socket currently being gated by the RN is compared with the desired class (as provided by the user) via the previously mentioned 4-neuron matching circuit (section \ref{sec:Classification}). This will only spike if the classes match, and is connected to the CoS neuron which, in turn, inhibits the intention and the precondition (terminating the first EB and releasing the next EB's intention).

The complimentary non-matching circuit tells us when the gated socket does not match that specified by the user input, and is connected to a CoD. This sends a single inhibiting spike to the selective DNF, causing an instability. The activation peak will collapse and, on account of the additional inhibition from the memory DNF, will form elsewhere. Hence the EB repeats with a new socket. The CoD is given a refractory period of 50 so as to avoid overly inhibiting the DNF beyond what is necessary (for quite some time after the DNF peak collapses spikes are still traveling through the select neuron field, classifier, and non-matching circuit).\\

\noindent \emph{EB 2 -- Visual Servoing}

The second EB, corresponding to the visual servoing, is comparatively simpler. The intention triggers the visual servoing agent, which then descends towards the identified socket guided by the selective DNF. Explicitly, the intention node is connected to all 5 directional neurons (section \ref{sec:VS}). Although they constantly receive current from the selective DNF via the visual servoing actor, the $v_{th}$ of these neurons is set such that they only spike once the intention node is activated (which, recall, occurs when the desired socket has been found).

The CoS awaits signal that we have made contact with the board, at which point the intention is inhibited and the final insertion (section \ref{sec:Insertion}) can be carried out). Again, the CoS consists of a single LIF Reset neuron.

A CoD is not deemed necessary here, yet in theory could be implemented to, for example, reset the servoing to the home position if the DNF peak is lost.

\section{RESULTS}
\label{sec:Results}

We first show the feasibility of running the entirety of the pipeline on real neuromorphic hardware. Pre-recorded event data is fed into the network and the components are individually unit tested with LAVA's built in probes to parameter tune and assure the behavior is as expected. Energy and latency are additionally measured. Then we design a simulated robotic arm in MuJoCo and deploy an identical network to test performance. Finally, we test the viability of the fundamental components of the network in a real world environment by running them on simulated LAVA in conjunction with a real robotic arm.

\subsection{Running on Loihi 2}
\label{sec:Loihi2}

\begin{figure*}[h]
\centering
\includegraphics[width=\textwidth]{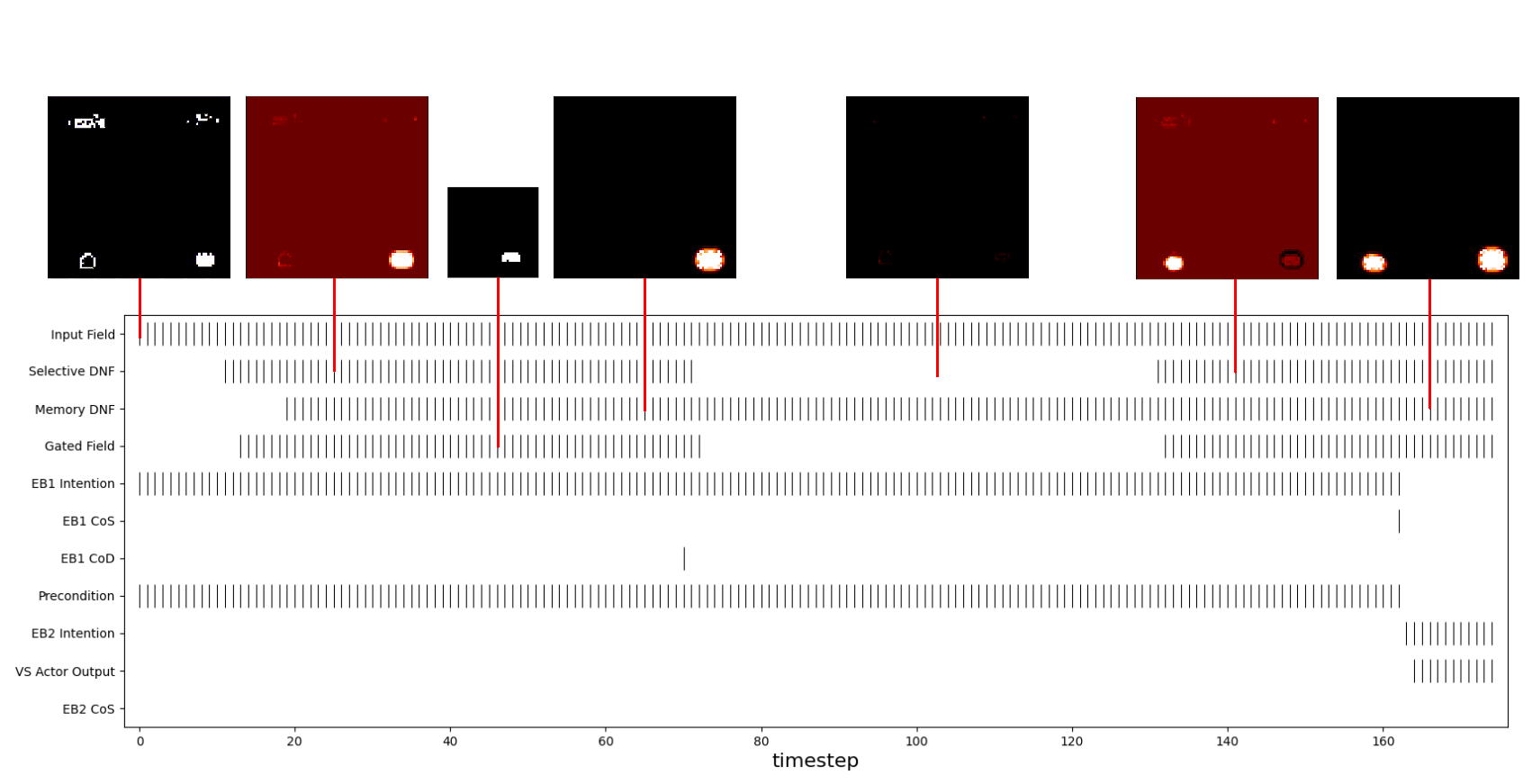}
\caption{An illustration of when various components within the architecture are seen to spike. Note that for neuronal fields such as the DNF we consider the layer to spike if at least one neuron is spiking. The inlaid images represent the voltages of the neuron fields for given timesteps. These are generated using LAVA voltage probes when running the network on real Loihi 2. Summary of inlays from left to right: event input, initial DNF selection, RN gated field, first peak stored in memory DNF, selective DNF peak collapsed by CoD (see single spike at timestep 70), new peak forms while previous socket is inhibited by memory DNF, new peak location is stored in memory DNF. Color scale is set such that black = minimum voltage (varies between fields and timesteps\textsuperscript{a}) and white $\geq v_{th}$ (spiking neurons). Note the EB2 CoS is designed to only spike when the plug makes contact with the socketboard. As here the input is pre-recorded and the output is not connected to the robot, this is not possible, and hence there is no spike.}
\scriptsize\textsuperscript{a} Note that for visualization purposes the black regions of the input field, gated field, and memory DNF represent zero; yet for the selective DNF, which can have strongly negative baselines, black represents $\sim-1e4$. The burgundy backgrounds in the $2^{nd}$ and $6^{th}$ inlays are still strongly negative ($\sim-7e3$).
\label{fig:raster_plot}
\end{figure*}

As multi-component neuronal architectures are large and complex networks involving many different neuron layers and connection types, it has not been immediately obvious that they can be handled by the current generation of neuromorphic hardware. Our pipeline is tested on multiple Intel Loihi 2 research chips \cite{Orchard21}. Figure \ref{fig:raster_plot} shows when various sections of the network spike relative to one another when run on-chip.

A single Loihi 2 chip contains 128 neurocores, each of which can simulate 8192 neurons in parallel. Our pipeline requires roughly 30,000 neurons. In theory these could be fit onto 4 neurocores. Yet the architecture is heavy in terms of connectivity (especially the two DNF layers). The synapses also require space on the chip, and therefore in practice we can put significantly fewer neurons on each neurocore. The compiler does not always partition the network in exactly the same manner from one run to the next (discussed further in section \ref{sec:discussion}). As a result, there is a slight variance in the number of neurocores used. However, in general the architecture occupies 2--3 chips and 250--300 neurocores.

A single inference takes 88~ms on the neuromorphic cores and consumes a mean of 78~mJ of energy (of which 59~mJ is dynamic energy\footnote{Defined as energy expenditure after removing the background energy cost of running the hardware.}). Total power is 0.88~W (of which 0.67~W is dynamic power). Table \ref{tab:energy_latency} shows a breakdown of energy consumption and latency.

After data input the first selective DNF neurons begin to spike 0.96~s and the peak has fully formed by around 1.8~s. In general, classification network outputs spikes by 5--6 seconds. If the socket is incorrect this is when the CoD destabilizes the DNF peak. A new peak takes around 4--5 seconds to form. Assuming this is the correct one, once this is subsequently classified and the CoS activates (another 3--4 seconds) the subsequent processes (precondition node inhibiting, EB2 intention node activation, visual servoing actor output) all happen effectively instantaneously (within 3 timesteps). At this point, the actor output can be read and validated, but as we are not connected to a real robot there is no subsequent step.

The actor output has the form $[-z,-y,+y,+x,-x]$. We check that when the USB is sought (located in the bottom-right of the FoV) the actor indeed returns [0,1,0,1,0], corresponding to a translation in the $-y$ and $+x$ directions (i.e. towards the bottom-right). If we instead search for the Ethernet socket, the output is [0,1,0,0,1] (i.e. towards the bottom-left) as expected.

\subsection{Simulated Robot}
\label{sec:sim}

\begin{figure}
\centering
\begin{subfigure}[h]{0.5\linewidth}
    \centering
    \includegraphics[width=0.97\linewidth]{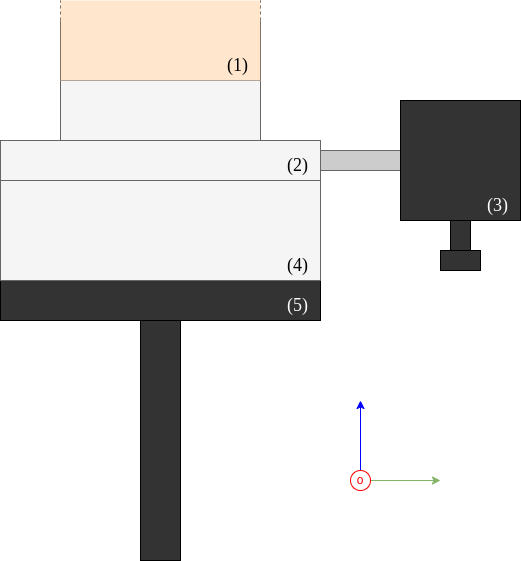}
    \subcaption{}
    \label{fig:camera_setup}
\end{subfigure}%
\begin{subfigure}[h]{0.5\linewidth}
    \centering
    \includegraphics[width=0.87\linewidth]{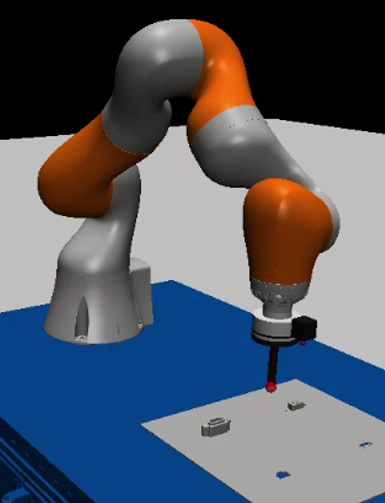}
    \subcaption{}
    \label{fig:sim}
\end{subfigure}
\caption{(a) Schema of end-effector with (1) KUKA arm, (2) metal plate and offset screw, (3) DVS camera, (4) force-torque sensor, and (5) peg with Ethernet plug. (b) Simulated arm and socketboard used to test the NSM. Based on \cite{Palinauskas23} with updated modeling to match our lab robot.}
\end{figure}

An identical architecture is tested within simulation using MuJoCo. Due to the inefficiency and high latency of connecting the simulation to the real Loihi 2 chip within the Intel vlab we instead opt for running the architecture in simulated LAVA. We use the simulation environment presented in \cite{Palinauskas23}, however the arm is refined to better match our real arm in the lab (addition of force torque sensor, Ethernet plug attached to the end-effector, and neuromorphic camera, all with realistic mass and MoI) and the socketboard is updated to a four-socket configuration. The camera is mounted to the arm with a slight offset, with the vision axis parallel to the end-effector peg. The focal point is adjusted such that the socket board fits perfectly into the FoV, and the offset is set to the minimum distance for which the plug does not obscure the FoV. See schema in Figure \ref{fig:camera_setup}. The robot's initial position is slightly varied between trials, and the vibrations are created by applying small random forces to the end-effector in the $xy$-plane. The simulation setup can be seen in Figure \ref{fig:sim}. An annotated video showing a full simulated insertion can be viewed at \cite{eleanor_link_sim}. %\href{https://www.youtube.com/watch?v=eJmGT4JlW5w}{https://www.youtube.com/watch?v=eJmGT4JlW5w}.

We apply the same downsampling as described in section \ref{sec:Data_Capture}. Although the I/O constraints do not exist when using simulated LAVA, as we are using here, imposing them gives a more realistic picture of the expected performance on real hardware.

The robot completes visual servoing to the correct socket 68\% of the time across 250 trials, where a trial is considered a success if the end-effector makes contact with the socket perimeter (the final force-torque guided insertion can then feel the socket edge and complete the insertion in the real robot --- see section \ref{sec:real} --- even if this is not implemented in simulation due to the complexity of perfectly modeling the collision profiles of each socket). Even when the desired socket is correctly identified the visual servoing can still fail due to DNF peak instabilities (due to rapid movement or unfavorable camera angles). In this case the end-effector can drift outside the socketboard, or impact a region of the socketboard where all sockets are beyond the camera's FoV. Regarding the classification accuracy, see a comparison of the simulated and real performance in section \ref{sec:Classification_Accuracy}.

Recall that although multiple teams have achieved visual servoing tasks with perfect or near-perfect success rates \cite{Muthusamy21, Ayyad23, Lawson23}, these are accomplished without a neuromorphic chip. The above-mentioned resolution constraints are therefore not present, and powerful classical CV techniques can be employed. Although \cite{Amaya24} does accomplish the insertion task with a neuromorphic chip and 100\% success, this is a simplified insertion with no classification, no visual servoing, and the run is initialized with the peg near the hole.

\subsection{Real Robot}
\label{sec:real}

\begin{figure}[h]
\centering
\includegraphics[width=0.3\textwidth]{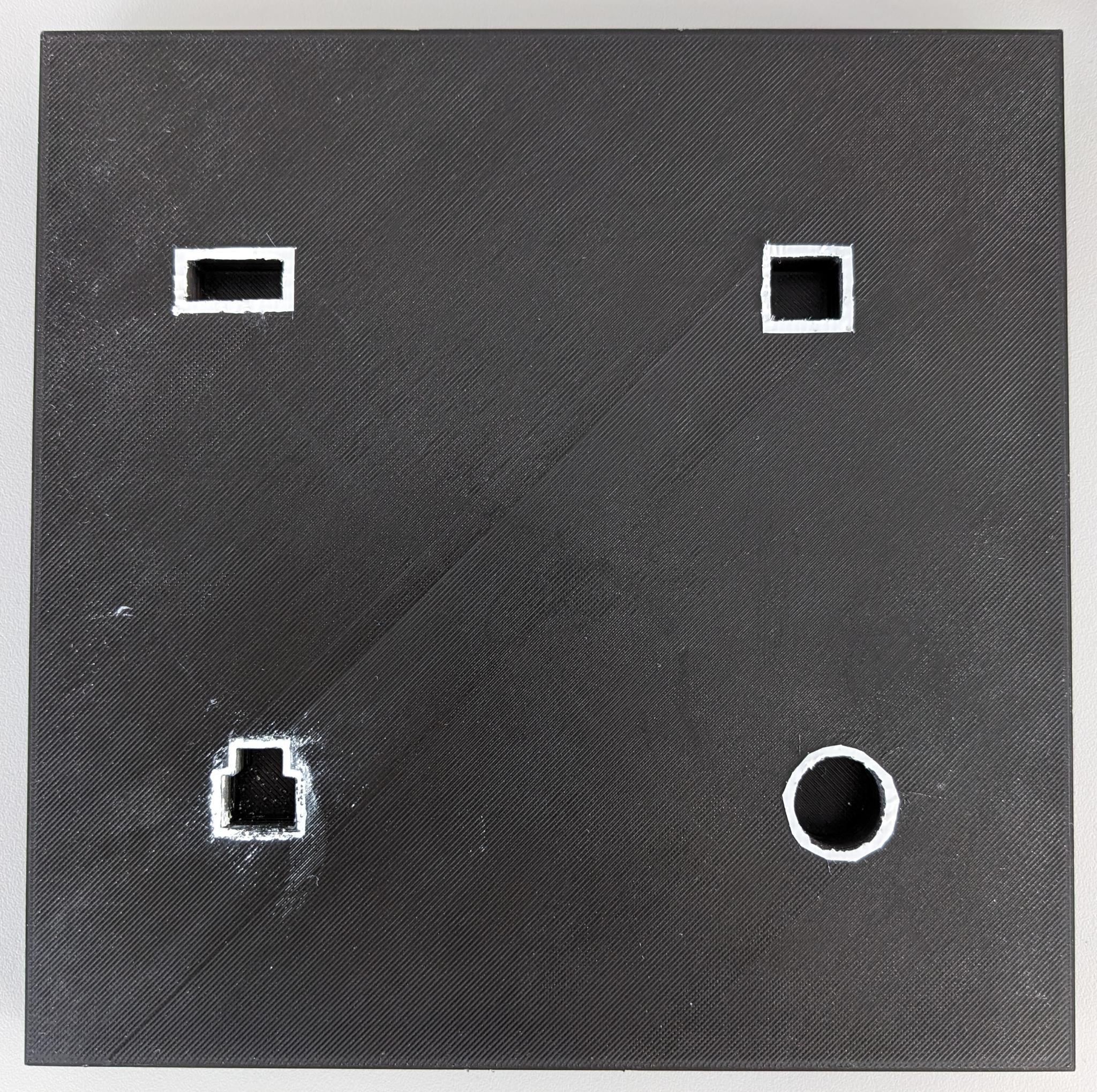}
\caption{3D printed socketboard. A white outline is added to the edge of the sockets to increase contrast, and therefore also event generation.}
\label{fig:socketboard}
\end{figure}

\begin{figure}[h]
\centering
\includegraphics[width=0.45\textwidth]{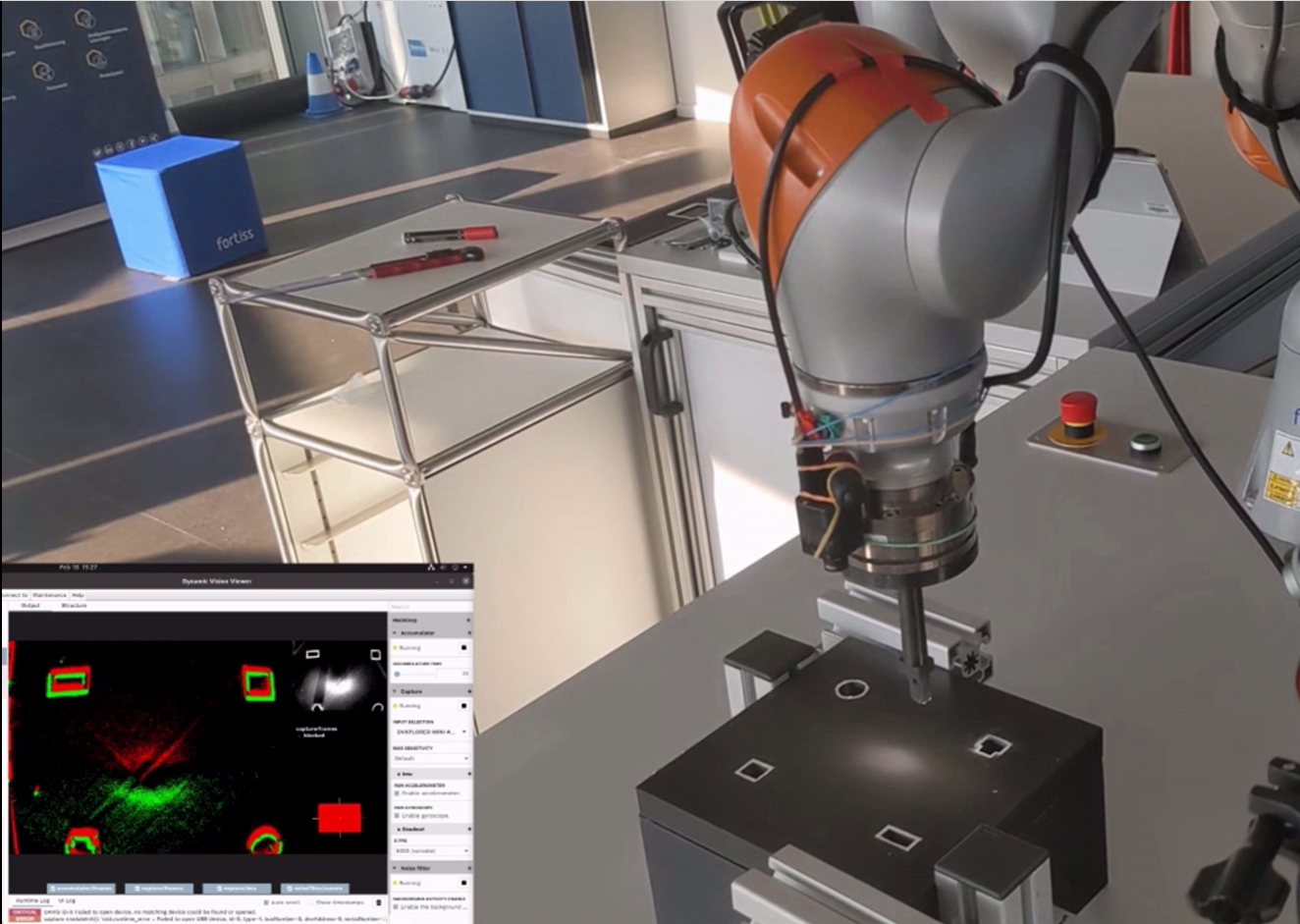}
\caption{Real setup showing the Ethernet plug, mounted neuromorphic camera, and vibrator. The display shows the raw events as seen by the camera (before downsampling).}
\label{fig:real}
\end{figure}

Finally, we test the pipeline on a real robotic arm, again using simulated LAVA for the reasons mentioned previously. Due to resource constraints, we test only the key components of the pipeline: the selective DNF, the classification system, the visual servoing actor, and the force-torque insertion. We build upon the setup presented in \cite{Amaya24} --- a 7DoF KUKA iiwa robot\footnote{Control the nullspace is not deemed relevant to the task, and therefore the 3$^{rd}$ joint is kept fixed, effectively resulting in 6DoF.}. An end-effector consisting of a peg with an Ethernet plug on the tip is 3D printed, as is a socketboard\footnote{For the real setup we use a slightly modified socketboard containing idealized sockets with 2~mm tolerances on each side. This was found to be necessary as the final insertion risked breaking the relatively thin plug.} (see Figure \ref{fig:socketboard}). An Inivation DVXplorer Micro neuromorphic camera is mounted in a similar manner to that described in section \ref{sec:sim} (see Figure \ref{fig:camera_setup}). For better event generation, a small outline of white paint is added around the sockets and spot lights are mounted on the robot bench, facing the middle of the socketboard. The full setup can be seen in Figure \ref{fig:real}.

The DNF works as expected, forming in roughly the same time as was seen on-chip (section \ref{sec:Loihi2}). In the real setup the classification is found to be perfectly accurate, even with downsampling (see caveats in section \ref{sec:Classification_Accuracy}). As we do not test the RN on the real robot\footnote{This is a planned future step.}, non-neural cropping is used here to hand segments to the classifier. When the correct socket is identified, the visual servoing works well, and the plug makes contact with the table with a mean offset of 4~mm. We do not witness the step-wise descent described in section \ref{sec:VS}, meaning the peg-camera offset is negligible within the FoV. The final force torque insertion consistently guides the plug in successfully. From home position to insertion is roughly one minute --- allowing the visual servoing to progress faster we begin to lose the DNF peak. This would likely be fixed with higher resolution.

An annotated video of the insertion on the real robot can be found here: \rev{\cite{eleanor_link}}. %\href{https://www.youtube.com/watch?v=A9D3PLCWQgQ}{youtube.com/watch?v=A9D3PLCWQgQ}. 
Note that, in this video, as the classification was found to have near-perfect accuracy, we manually force an initial misclassification and subsequent masking. Besides this, the classification itself, selective DNF, and visual servoing are run on simulated LAVA, and the force-torque insertion is run on real Loihi 1 (due to resource constraints it was not possible to move the network to Loihi 2).

\subsection{Classification}
\label{sec:Classification_Accuracy}

\begin{figure}
\centering
\includegraphics[width=0.4\textwidth]{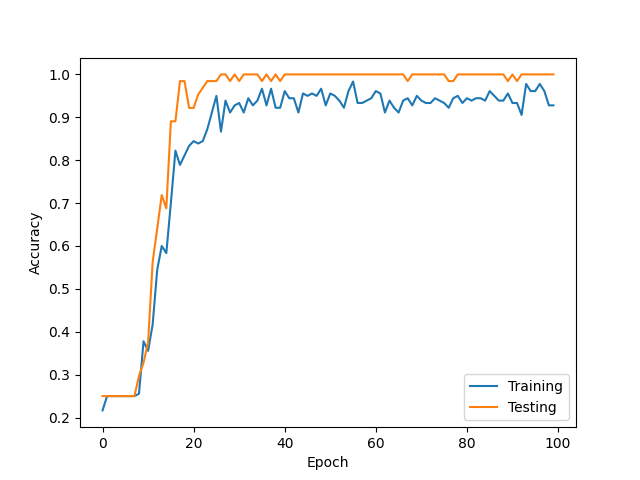}
\caption{Classification training accuracy measures for real event data on the real robot (see section \ref{sec:Classification} for an overview). As the positions and orientations of the sockets are always fixed, data augmentation is only applied to the train data, hence why the test accuracy is consistently higher.}
\label{fig:accuracy}
\end{figure}

The classification is always carried out from a pre-defined home position. This means that from one run to the next the sockets are in exactly the same position and orientation. Therefore, the classification is relatively trivial. Indeed, Figure \ref{fig:accuracy} shows the training on real event data generated while the camera is vibrating in the home position. Note that the training accuracy is lower than the test accuracy because we include data augmentation only for the training data. Test accuracy is nearly always 100\%. 

In simulation, the classification accuracy is lower than for the real robot, roughly 90\%. This is because the inclusion of the RN, which combines the selective DNF with the input in order to gate a region (section \ref{sec:RN}), can have the effect of obscuring some of the edges of the socket. Ideally the activation peak in the DNF should be marginally larger than the socket, yet through manual parameter tuning this is difficult to achieve in practice, especially when the sockets are slightly different sizes. Another consideration is that, unlike the real setup which uses simplified idealized sockets (for reasons mentioned in section \ref{sec:real}), the sockets in the simulation are true industrial sockets with a large amount of structure difficult to capture with downsampled resolution.

\subsection{Results Summary}
\label{sec:results_summary}

\centering
\captionof{table}{Simulation and Real Performance} \label{tab:SimReal} 
\begin{tabular}{l|c|c}
& Simulated & Real\\
\hline
Individual Classification Acc. & 90\% & 100\%* \\
Successful Visual Servoing & 68\% & 100\% \\
Mean Offset & 5~mm & 4~mm\\
Successful Insertion & NA & 70\% \\
Average Insertion Time & $\sim$ 1 min & $\sim$ 1 min \\
\end{tabular}

{~~~~~~~~~~~~~~~~~~~~~~~~~~~~~~~~~~~~~~~~~~~~~~~~~~~~~~~\scriptsize*See \S\ref{sec:Classification_Accuracy}.}
\vspace{1cm}
\centering
\captionof{table}{Loihi 2 Profiler Values} \label{tab:ProfilerValues} 
\begin{tabular}{ccc|c|c}
\label{tab:energy_latency}
 & Power (W) & & Latency & Energy per\\
Static & Dynamic & Total & (ms) & Step (mJ)\\
\hline
0.22 & 0.67 & 0.88 & 88 & 59
\end{tabular}
\justifying
\vspace{1cm}

\section{Discussion}
\label{sec:discussion}

\subsection{On-Chip Performance}

Our pipeline shows the viability of incorporating multiple neuronal components entirely on-chip. Notably, we find power can remain in the milli Watt regime (0.67~W) even when including systems for attention, memory, gating (RN), classification, visual servoing, and overarching logic (NSM). Although an identical algorithm does not exist for GPU (in part because many of the components, such as the DNF, are neuron-native) we can still draw some comparisons. For example, a linear pipeline for line detection consumes $\sim5$~W on a mobile GPU platform\footnote{Note that in the cited work the minimum resolution is $320\times180$, larger than our $80\times80$. However, figures 10 \& 11 in \cite{Suder23} show that, in fact, the effect of resolution on overall power consumption is minimal, and most of the consumption is from the subsequent data processing steps.} \cite{Suder23}. It is therefore safe to assume that a pipeline of our complexity deployed on GPUs would run in \emph{at least} the tens-of-watts regime. Although in or case not possible due to I/O constraints, directly sending events from the DVS to the chip would only add a few tens-of-milli Watts to the system \cite{Gallego22}.

Our mean per-timestep latency value of 88~ms is on par with other edge applications. For example, in \cite{Bueno25} YOLO v8-s is used for classification on an edge device (Raspberry Pi 5 interfaced with the Hailo-8L AI accelerator) and the measured per-frame inference latency is 53~ms. Again, recall that this is classification only --- incorporating all other components of our system would push the latency much higher. Another point is that for each timestep there is currently the previously mentioned bottleneck for spike injection and extraction (even when data is loaded from a pre-recorded file). The packet processing times would likely be much smaller without this, as is expected to be the case in future generations of hardware.

As per the overall functioning of the components on real Loihi~2, it has been shown that they can operate together for the stated robotics application. However, testing such a large and heterogeneous architecture reveals some challenges. There remains some randomness is the resource allocation process intrinsic within the LAVA compiler. That is to say, an identical code run multiple times on-chip can have slightly different neurocore partitioning. For most SNN, layers can be easily divided. Yet our pipeline, lacking clearly defined layers, does not have a single `natural' way of partitioning (consider, for example, how the selective DNF is connected to 6 other components). There will be small differences in spike timings between partitions, and while for most SNN this would be negligible, we do see its effects in our network. DNF peaks form at slightly different timesteps, and as such downstream processes can also be affected. In the same vein, connecting probes or modifying a single parameter can lead to non-negligible changes in the behavior of the system.

Along a similar line, across the tens of thousands of neurons and synapses in our pipeline there are 104 tuneables parameters\footnote{This figure includes only the relevant parameters for our purposes within the pipeline, and excludes those related to I/O or general overhead chip management. For a given neuron field, all the LIF neurons are set to have the same parameter values (this is also true for all the weights between two fields, with the exception of those within the RN and visual servoing actor), although in general this doesn't need to be the case.} that are relevant to the behavior (mainly the $du$, $dv$, and $v_{th}$ terms from equations \ref{eq:u} and \ref{eq:v}, as well as the synapse weights). These were initialized based on what prior examples were available, as well as mathematical estimation, however it was almost always necessary to subsequently adjust them through trial and error in order to fine-tune the behavior: a tedious process. Nowhere was this more true than with the selective DNF, whose connections to many other components resulted in an interplay between 34 tunable parameters. If not set correctly, a new peak would not form after the collapse of a previous one, unwanted secondary peaks would form, a single peak would sometimes grow to swallow the entire field, etc. If multi-component pipelines such as this are to be viable, manual engineering will need to be replaced by a system of self-regulation. Something comparable has been accomplished within a simple 200-neuron network by \cite{Maryada25} on DYNAP-SE2 using homeostasis for regulating runaway excitation. Although Loihi 2 does support homeostasis at the neuron-level, exploring this in the context of a large 6400 neuron DNF, especially when there is the added complexity of memory DNF inhibition and NSM integration, is well beyond the scope of this project.

%\subsection*{Innatera Pulsar}

\rev{A final noteworthy chip-related point worth mentioning pertains to the 2025 release of the Innatera Pulsar neuromorphic micro-controller \cite{innatera_pulsar}, referred to as a `Spiking Neural Processor' (SNP). The Pulsar SNP contains a heterogeneous architecture including both analog and digital neurons, as well as traditional non-spiking CNN accelerators. The SNP is advertised as handling multiple data types (audio, motion, vibration, etc.), yet it appears to be more targeted towards one-dimensional sensor input for now. The corresponding Talamo SDK does not appear to have out-of-box methods for naturally integrating multimodal data into a single network, nor for neuron-native process orchestration (ie. DNF-bsed methods such as ours) --- these would likely have to be built from scratch as we have done herein with LAVA. Nonetheless, the SNP would allow for direct on-chip parallelization between neuronal and non-neuronal networks. Even if encoding would still be necessary, this could prove hugely advantageous over resorting to off-chip processes. Ultimately, and not in the least because Talamo is not open-source, it is simply too early to know exactly what the Innatera Pulsar may be capable of.}

\subsection{Robotic Implications}
Time and resource constraints prevented a full chip-robot integration. Yet it was still possible to test the pipeline in simulated LAVA, guiding a robotic arm in MuJoCo. The primary components of the network were also run in simulated LAVA connected to a robot arm. In a sense, these represent a `best case' scenario for what could be expected with full chip-robot integration (as off-chip we are not subject to on-chip partition considerations, and can more easily parameter tune).

In simulation, we learn that the size of the DNF peak is crucial for the RN to maintain the full details of the socket events when gating (section \ref{sec:Classification_Accuracy}). Addressing this is non-trivial, as sockets have different sizes, shapes, and density of internal structure. The USB socket, which has a lot of internal structure in the form of a wide central pin with metallic strips, generates lots of events and is found to consistently create the first selective DNF peak. By contrast, the Ethernet socket is effectively a simple square hole with narrow edges and no internal structure; as a consequence, very few events are generated. The selective and memory DNFs must be parametrized such that peaks are not overly excitatory lest the USB cause runaway growth, yet also sufficiently strong such that the peak fully covers the Ethernet. The difficulty of achieving this balance is responsible for the slightly reduced classification performance in simulation (Table \ref{tab:SimReal}).

One potential way to address this could be a pre-processing layer between the event intake and selective DNF. For example, applying proto-object saliency \cite{D’Angelo22} could return regions of normalized excitation wherever a proto-object (a socket in our case) is located, regardless of event density.

On the real chip, it is not possible to measure accuracy as the input data is pre-recorded and therefore, for a given partition, the peak formation and subsequent classification will be the same for each run. Nonetheless, this first foray of testing the pipeline on the real chip suggests the on-chip accuracy would likely be worse --- the outer-edge clipping caused by the RN seems qualitatively more severe than with simulated LAVA, and for many partitioning configurations and parametrizations the Ethernet is misclassified (Figure \ref{fig:raster_plot} shows a successful partition picked among numerous that misclassified).

We can also evaluate the success of the DNF-based actor in guiding the visual servoing on the real robot. The average offset from the socket center when impacting the socketboard is 4~mm. Contrasting this to previous neuromorphic visual servoing efforts (all using a DVS integrated with non-neuromorphic hardware) we have 2--3~mm in \cite{Ayyad23}, 10--24~mm in \cite{Muthusamy21} and 34~mm in \cite{Lawson23}. For classical robots/cameras, \cite{Sileo24} reports table-impact offsets from 2--5~mm. This should serve to validate the DNF as a reasonable method for visual servoing.

\begin{figure}[h]
\centering
\includegraphics[width=0.45\textwidth]{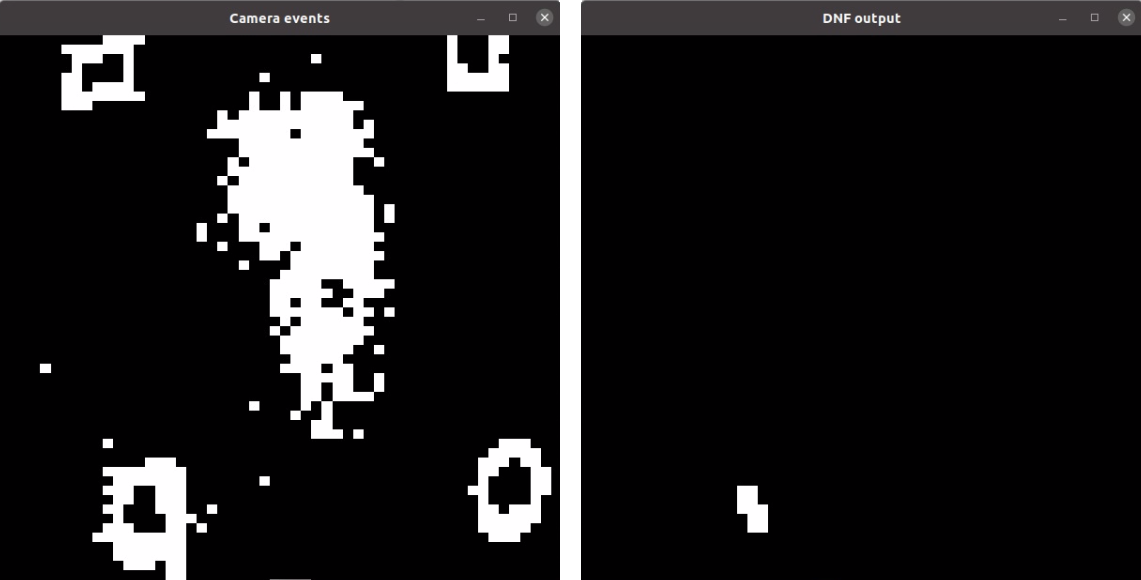}
\caption{Left: A frame of input events (downsampled to $80\times80$) during the visual servoing showing intense glare on the reflective surface of the socketboard. Right: the DNF peak remains on the bottom left socket.}
\label{fig:glare}
\end{figure}

The DNF also provides convenient resistance to noise. The 3D printing material is a plastic filament which leaves the socketboard (Figure \ref{fig:socketboard}) mildly reflective. As a result, for some frames during the visual servoing, bright glare can appear in the FoV on account of the bright spot lights required for sufficient event generation (Figure \ref{fig:glare}). Yet note how, on the right of Figure \ref{fig:glare}, the DNF peak remains locked on the selected socket. We expect this robustness to perturbation would continue to exist with real chip integration, as when running on-chip we find the DNF peak --- although sometimes delicate during formation --- tracks well once fully formed.

As for the final insertion once the plug has made contact with the table, recall this is not tested for the simulated robot to avoid having to create full collision models for the sockets. On the real robot, the insertion is successful in 70\% of trials (Table \ref{tab:SimReal}), where a failure is when the peg is not inserted within 30 seconds of making contact with the table. The 3D printed plug attached to the end-effector is quite delicate, requiring us to drastically limit the maximum force in the $-z$ direction to avoid damage. On account of this, the peg is sometimes seen to `float' over the hole while searching without enough downwards force to complete the insertion (contrast this with \cite{Amaya24} where the wider peg does not experience this issue). Other sources of failure include the peg getting `stuck' on the painted white edge of the socket (recall our goal is not perfect friction modeling), or the DNF peak forming on a patch of glare present at the very beginning of the servoing (mistaking it for a socket).

The vibrations also cause complications relating to the DVS camera. They must have sufficient amplitude to vibrate the camera, but not too large as to vibrate (and damage) the actuators in the robotic arm. At the same time, the camera must remain pointed at the table. Even a tiny tilt will shift the FoV and result in an offset between the socket and plug upon impact. We attach the camera with a flexible support system such that it can still vibrate (previously a screw attachment was found to be too rigid for proper event generation). Often, the vibrations would induce the above-mentioned tilting, or uncalibrate the lens (for which rotations alter the focal depth). For future work, ideal would be some form of `cage' attached solidly to the robotic arm, within which the camera could vibrate without tilting.

\subsection{Outlook and Conclusions}

As with biological systems, robust behavior in neuromorphic robots will require multi-component systems of heterogeneous neuronal architecture such as that presented herein. This is necessary not only for more complex computation, but also for multimodal data integration. Our pipeline is a first attempt at such a solution. It modestly links two elementary behaviors (searching and visual servoing) consisting of six components using neuron-native orchestrational logic.

Expanding to include more components and behaviors, our methodology could serve as a foundation. What is now needed is a more robust system for neurons, both individually and within groups, to self-regulate their parameters. It is unrealistic to continue finding the balance between runaway excitation and insufficient activation through tedious hand-tuning. This could potentially be remedied through homeostasis, or some facet of on-chip learning. Both avenues represent fascinating directions for future research, especially in the context of expanding upon DNF.

The relational network, which offers a method for on-chip covert attention through gating, also merits further exploration. Notably, layering multiple weight matrices could allow for one among multiple overlapping regions and scales to be gated and handed onward to a classifier on-chip. 

With proper chip integration on the real robot, the robustness of the pipeline could be better tested in the face of perturbation. For example, it remains to be seen whether glare (Figure \ref{fig:glare}) would cause issues for an on-chip DNF, and to what degree the edge-artifacts mentioned in \S \ref{sec:Classification_Accuracy} would hinder classification.

Nonetheless, our first on-chip tests have shown operational power remains in the sub-milli Watt regime, and latency is competitive with the state-of-the-art. The NSM also succeeds at offering the structure needed to administer the multiple pipeline components, as we had hoped. This provides encouragement that complex neurorobot behavior is indeed within reach, and should act as a rallying call within the neuromorphic community to double down our efforts.

\section*{Code}
\begin{center}
Code can be found at \href{https://github.com/EvanEames/NSM}{https://github.com/EvanEames/NSM}
\end{center}

\section*{Acknowledgements}

We offer thanks to Mathis Richter from Intel Labs. He contributed immensely in the forms of guiding pipeline development, sharing wisdom on the various components, and countless sessions of LAVA code debugging. We would also like to thank Yulia Sandamirskaya, who played an early part in conceiving of the project and getting the groundwork set up while still with Intel.\\

This research was funded by the Bavarian Ministry of Economic Affairs, Regional Development and Energy (StMWi). Funding code DIK-2105-0036// DIK0368/01.\\

The authors have no conflicts of interest to declare.

\onecolumn
\centering
\begin{appendices}
\section{Pipeline Parameters}
\label{sec:Parameters}

\centering
\captionof{table}{Neuron Parameters} \label{tab:NeuronParams} 
\begin{tabular}{c|c}
\hline
Selective DNF (section \ref{sec:DNF})\\
\hline
LIF: $du, dv, vth$ & 809, 2047, 30\\
Kernel values: & \\
$amp\textrm{\_}exc$ & 6\\
$width\textrm{\_}exc$ & [12,12]\\
$amp\textrm{\_}inh$ & 0\\
$width\textrm{\_}inh$ & [1,1]\\
\hline
Memory DNF (section \ref{sec:Memory})\\
\hline
LIF: $du, dv, vth$ & 2000, 2000, 30\\
Kernel values: & \\
$amp\textrm{\_}exc$ & 13\\
$width\textrm{\_}exc$ & [10,10]\\
$amp\textrm{\_}inh$ & -10\\
$width\textrm{\_}inh$ & [12,12] \\
\hline
Intermediary Neuron (section \ref{sec:DNF})\\
\hline
LIF: $du, dv, vth$ & 4095, 3300, 10\\
\hline
RN Select Neuron Field (section \ref{sec:RN})\\
\hline
LIF Reset: $du, dv, vth$ & 1000, 3300, 3\\
\hline
RN Gated Region \\
\hline
LIF: $du, dv, vth$ & 300, 1000, 1\\
\hline
Classification (section \ref{sec:Classification}) \\
\hline
Input & $40\times40$\\
Conv 1 & $5\times5$, stride = 1, pad = 2 \\
Pool 1 & $2\times2$, stride = 2, pad = 0 \\
Conv 2 & $3\times3$, stride = 1, pad = 1 \\
Pool 2 & $2\times2$, stride = 2, pad = 0 \\
Dense 1 & $1\times1\times512$\\
Dense 2 & $1±\times1\times4$\\
Output & $4$\\
\hline
Matching/Non-Matching Circuits (section \ref{sec:Classification}) \\
\hline
Matching: & \\
LIFReset: $du, dv, vth$ &  4095, 3300, 3\\
Non-Matching: & \\
LIFReset: $du, dv, vth, bias\textrm{\_}mant, bias\textrm{\_}exp$ &  4095, 3300, 3, 1, 7\\
\hline
Visual Servoing Actor (section \ref{sec:VS})\\
\hline
LIFReset: $du, dv, vth$ & 4095, 3300, 62\\
\hline
NSM Components (section \ref{sec:Full Architecture})\\
\hline
EB1 Intention & \\
LIF: $du, dv, vth, bias\textrm{\_}mant, bias\textrm{\_}exp$ & 4095, 3300, 1, 1, 7\\
EB1 CoS & \\
LIF Reset: $du, dv, vth$ & 4095, 3300, 3\\
EB1 CoD & \\
LIF Refractory: $du, dv, vth, refractory\textrm{\_}delay$ & 4095, 3300, 1, 50\\
Precondition & \\
LIF: $du, dv, vth, bias\textrm{\_}mant, bias\textrm{\_}exp$ & 1000, 1000, 1, 1, 7\\
EB2 Intention & \\
LIF: $du, dv, vth, bias\textrm{\_}mant, bias\textrm{\_}exp$ & 1000, 1000, 1, 1, 7\\
EB2 CoS & \\
LIF Reset: $du, dv, vth$ & 4095, 3300, 3\\
\hline
\end{tabular}
\newpage

\centering
\captionof{table}{Weight Values} \label{tab:WeightParams} 
\begin{tabular}{c|c}
\hline
Input to Selective DNF & 8\\
Selective DNF to Intermediate Neuron & 1\\
Intermediate Neuron to Selective DNF & -10\\
Selective DNF to Memory DNF & 8\\
Memory DNF to Selective DNF & -5\\
Input to Select Neuron Field & 2\\
Selective DNF to Select Neuron Field & 2\\
Select Neuron Fielt to Gated Region & 2\\
Gated Region to Classifier & 1\\
User Input to Matching \& Non-Matching Circuits & 2\\
Classifier Output to Matching \& Non-Matching Circuits & 2\\
Matching Circuit to EB1 CoS & 2\\
Non-Matching Circuit to EB1 CoD & 2\\
EB1 Intention to EB1 CoS & 2\\
EB1 Intention to Selective DNF & 0*\\
EB1 CoS to EB1 Intention & -100\\
EB1 CoD to Selective DNF & -120 \\
EB1 CoS to Pre-condition & -100\\
Pre-condition to EB2 Intention & -1\\
EB2 Intention to Visual Servoing Actor Output & 37\\
EB2 Intention to EB2 CoS & 2\\
EB2 CoS to EB2 Intention & -100\\
Selective DNF to Visual Servoing Actor & 2\\
\hline
\end{tabular}

{\scriptsize * This could be used to initialize the DNF when the EB1 intention is first activated.

However, as our EB1 intention is automatically active from the beginning, it is not deemed necessary.}

\end{appendices}

\twocolumn
\newpage

%%%%%%%%%%%%%%%%%%%%%%%%%%%%%%%%%%%%%%%%%%%%%%%%%%%%%%%%%%%%%%%%%%%%%%%%%%%%%%%%

\bibliographystyle{unsrt}
\bibliography{Eleanor2025}

\end{document}